\pdfoutput=1

\documentclass[11pt]{article}

\usepackage{EMNLP2023}

\usepackage{times}
\usepackage{latexsym}

\usepackage[T1]{fontenc}

\usepackage[utf8]{inputenc}
\usepackage{kotex}

\usepackage{microtype}

\usepackage{inconsolata}

\usepackage{algorithm} %
\usepackage{algorithmic} %
\usepackage{amsmath} %
\usepackage{graphicx} %
\usepackage{tabularx} %
\usepackage{afterpage} %
\usepackage{caption} %

\usepackage{array}
\usepackage{booktabs}
\usepackage{multirow}
\usepackage{multicol}

\title{Expanding Search Space with Diverse Prompting Agents: An Efficient Sampling Approach for LLM Mathematical Reasoning}

\author{
 \textbf{Gisang Lee{\textsuperscript{1, 2}}},
 \textbf{Sangwoo Park\textsuperscript{1}},
 \textbf{Junyoung Park\textsuperscript{3}},
 \textbf{Andrew Chung\textsuperscript{3}}, \\
 \textbf{Sieun Park\textsuperscript{4}},
 \textbf{Yoonah Park\textsuperscript{3}},
 \textbf{Byungju Kim\textsuperscript{2}\footnote[2]{}},
 \textbf{Min-gyu Cho\textsuperscript{2}\footnote[2]{}}
\\
 \textsuperscript{1}KAIST,
 \textsuperscript{2}Mathpresso Inc.,
 \textsuperscript{3}Seoul National University, \\
 \textsuperscript{4}Goldsmiths, University of London
\\
   \texttt{\{bobopack, swgger\}@kaist.ac.kr},
   \texttt{\{jyp0314, aschung01, wisdomsword21\}@snu.ac.kr}, \\
   \texttt{sieunpark77@gmail.com}, 
   \texttt{\{peyton.kim, mike.cho\}@mathpresso.com}
}

\begin{document}
\maketitle
\begin{abstract}

Large Language Models (LLMs) have exhibited remarkable capabilities in many complex tasks including mathematical reasoning. However, traditional approaches heavily rely on ensuring self-consistency within single prompting method, which limits the exploration of diverse problem-solving strategies. This study addresses these limitations by performing an experimental analysis of distinct prompting methods within the domain of mathematical reasoning. Our findings demonstrate that each method explores a distinct search space, and this differentiation becomes more evident with increasing problem complexity.
To leverage this phenomenon, we applied efficient sampling process that uniformly combines samples from these diverse methods, which not only expands the maximum search space but achieves higher performance with fewer runs compared to single methods.
Especially, within the subset of difficult questions of MATH dataset named \textbf{MATH-\textit{hard}}, The maximum search space was achieved while utilizing approximately \textbf{43\%} fewer runs than single methods on average. These findings highlight the importance of integrating diverse problem-solving strategies to enhance the reasoning abilities of LLMs.

\end{abstract}

\section{Introduction}

Recent advancements in large language models (LLMs) have significantly enhanced their reasoning abilities, particularly in mathematical reasoning and code generation. High-performing models such as GPT-4o (\citealp{openaiblog}), Claude Opus (\citealp{claude}) have demonstrated their capabilities in these challenging domains, showcasing their advanced performance. These models are typically employed through step-by-step natural language reasoning methodologies named Chain-of-Thought (CoT) to ensure the validity and accuracy of their solutions (\citealp{wei2023chainofthought}). Particularly in solving math problems, existing approaches either focus on validating the logical sequence during the solution process (\citealp{zhang2024cumulative}; \citealp{wang2024rat}; \citealp{zhou2024dont}), seek verification support for complex calculations (\citealp{chen2023program}; \citealp{zhou2023solving}; \citealp{zhong2024debug}), or aim to secure both logic validation and calculation accuracy (\citealp{gou2024tora}). A common feature of these methods is the use of sampling and voting processes to achieve self-consistency (CoT-SC) by generating multiple solutions (\citealp{wang2023selfconsistency}).

\begin{figure}[t]
    \centering
    \includegraphics[width=0.5\textwidth,keepaspectratio]{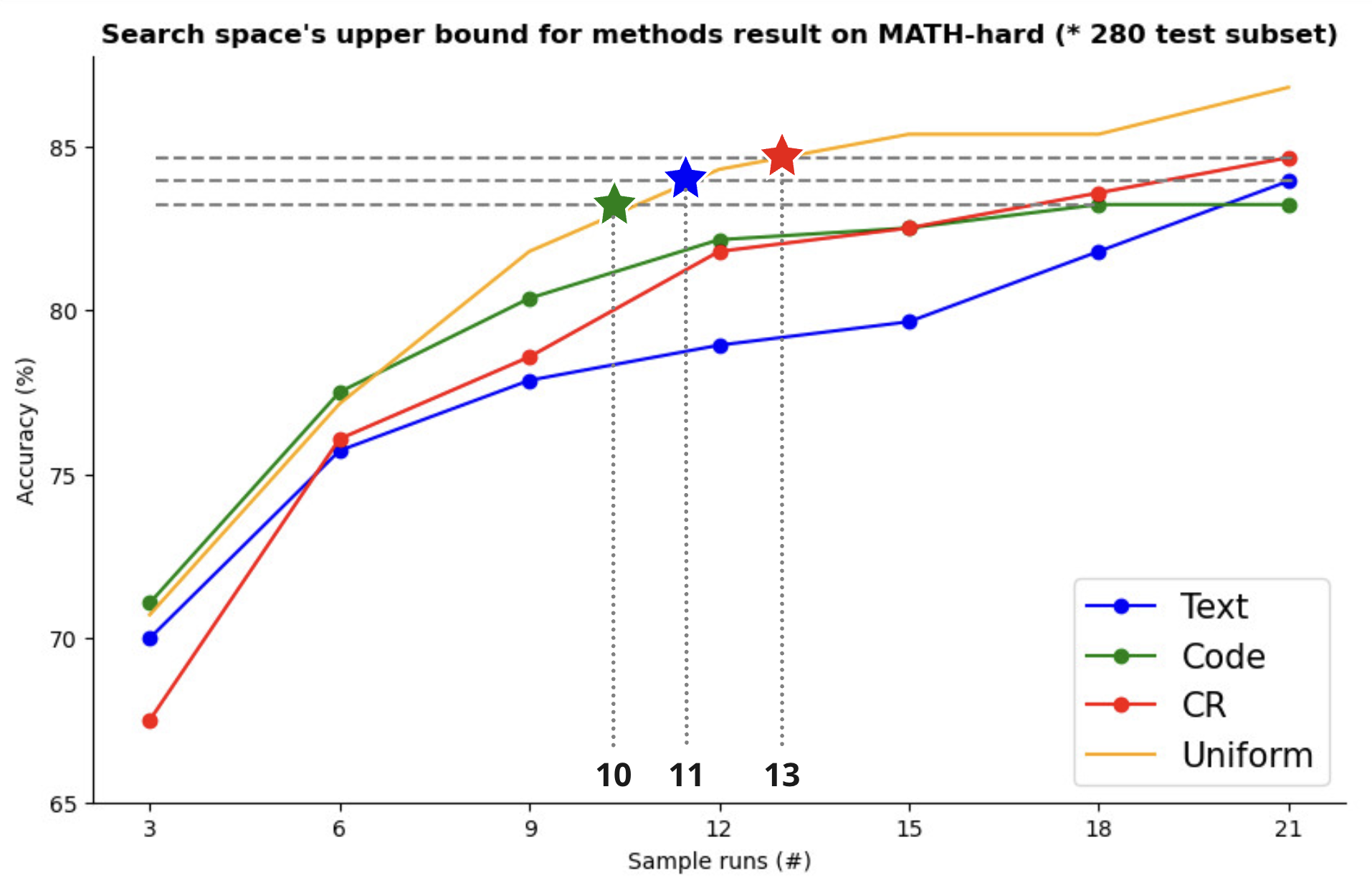}
    \caption{Line graph of maximum search space's accuracy achieved by sampling 21 runs per methods. The three grey horizontal lines represent the upper bound values within a single method. The star markers indicate the points at which these upper bound values were achieved using our proposed Uniform Sampling method. It can be observed that for \textit{text}, \textit{code}, and \textit{CR}, the same upper-bound was reached while utilizing approximately 48\%, 45\%, and 35\% fewer runs, respectively.
    }
    \label{fig:exp_1_1}
\end{figure}

While these methods have been effective in verifying the solutions provided by LLMs and enhancing their reliability, they heavily rely on temperature adjustments to increase the diversity of problem-solving approaches. This reliance on self-consistency within their own generated solutions limits their ability to explore a wider range of problem-solving strategies. In contrast, human problem solvers invest significant effort not only in verifying the validity and accuracy of their calculations but also in exploring many potential solutions.

Recent efforts in the field of LLM's high reasoning have focused on integrating diverse agentic problem-solving methods to address these limitations (\citealp{du2023improving}; \citealp{liu2023dynamic}). Although these studies have shown promising performance on benchmarks such as MATH (\citealp{hendrycksmath2021}) and GSM8K (\citealp{cobbe2021gsm8k}), they lack a comprehensive analysis of why different agents collectively achieve high performance. Furthermore, there is an absence of methodologies that explore how the unique characteristics of each approach can be effectively integrated, beyond merely improving performance metrics.

Therefore, this study aims to address these gaps by performing an experimental analysis of the problem-solving strategies employed by various LLM agents within the domain of mathematical reasoning. Furthermore, we propose an efficient sampling process to effectively combine these diverse agents. Key observations obtained by experimental analysis and our contributions are as follows.

\paragraph{Observation}

To specifically evaluate the high reasoning abilities of LLMs, we analyzed state-of-the-art methodologies on the MATH dataset, which requires higher capabilities than GSM-8K. We categorized the approaches into three main prompting methods: \textit{Text}, \textit{Code}, and \textit{CR} (Cumulative Reasoning). We discovered that each method explores a distinct search space when generating correct answers, and this differentiation becomes more evident as the complexity of the problems increases.

\paragraph{Contribution}

We observed that each prompting method explores a different search space, and this realization led us to an efficient sampling strategy. Instead of inefficiently generating multiple samples within a single method, we demonstrated that uniformly mixing samples from these distinct methods significantly increases the maximum search space. As shown in Figure~\ref{fig:exp_1_1}, within the MATH-\textit{hard} subset, the maximum search space was achieved while utilizing approximately 43\% fewer runs than single methods on average.

\section{Method}

\begin{figure}[t]
    \centering
    \includegraphics[width=0.5\textwidth,keepaspectratio]{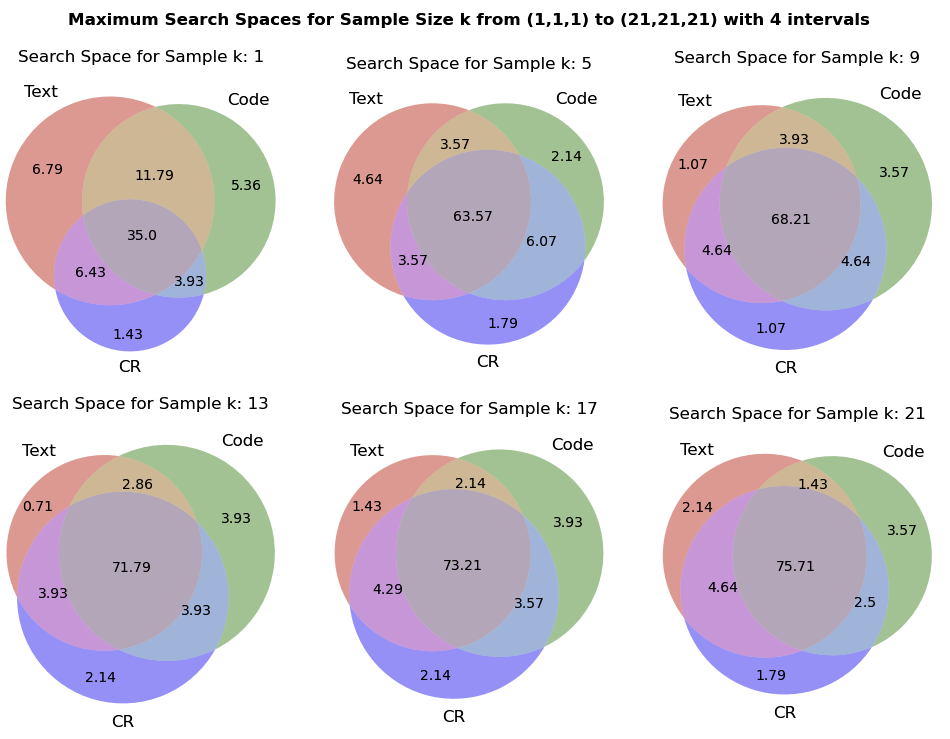} %
    \caption{Maximum search space for methods result on MATH-
\textit{hard} (* 280 test subset). From above, the Venn diagram's \( B \cup C \ - A \) represents the proportion of the search space that method A fails to explore.}
    \label{fig:output_space}
\end{figure}

\subsection{Expanding search space}

Figure~\ref{fig:output_space} shows a Venn diagram visualizing the maximum search space within the MATH-hard problems for the three prompting methods. We increased the sample sizes sequentially from (1,1,1) to (5,5,5) in intervals of 4, and finally up to (21,21,21) to see if this phenomenon persisted. The results showed that as the sample sizes increased, the overlap in the center gray area, representing the shared search space, grew. Although the absolute size of each unique search space decreased, the proportion of the search space that any single method (Method A) could not explore \( B \cup C \ - A\) remained within a certain bound. This demonstrated that even as the sample size k increased, the search spaces of each method remained robustly distinct.

\paragraph{Prompting methods}

We selected three prompting methods to analyze the differences in problem-solving approaches within the MATH dataset, building on the assumption that each method explores a distinct search space. We chose the following three prompting methods: (1) \textit{Text}, (2) \textit{Code} and (3) \textit{CR}.
\begin{enumerate}
\item \textit{Text}: As reported in \citealp{wei2023chainofthought}, this prompting method encourages natural language explanations using the Chain-of-Thought (CoT) approach. This serves as the base reasoning method of LLMs. The token cost for CoT-SC is used as the baseline.
\item \textit{Code}: This method directs the model to extract and execute code to derive the answer. Inspired by \citealp{chen2023program}, we specifically adopted the prompt presented in \citealp{gou2024tora}, characterized by converting natural language problems into local code interpreter. According to the average of the logged values in our experiments, the token cost for \textit{Code} is 3.0 times higher than the base text method.
\item \textit{Cumulative Reasoning (CR)}: The \textit{CR} framework, proposed by \citealp{zhang2024cumulative}, utilizes multiple LLMs cumulatively and iteratively for mathematical reasoning, mirroring human thought processes for problem-solving. We used \textit{CR with code} to remove additional environmental variables besides the prompts aspects when comparing with \textit{Code (Method 2)}.
\end{enumerate}

\paragraph{Selecting (Sampling)}

Although we secured a pool of runs by generating $n$ runs from each method, achieving an advantage in exploration over CoT-SC from a single method requires that the search space covered by these runs is extensive. Therefore, selecting a fixed number of runs should ensure high accuracy. To achieve this, an appropriate sampling algorithm that can effectively and efficiently combine solutions from various methods is necessary. To ensure that the final selected runs are as diverse as possible, we employed a method called uniform sampling.

\begin{quote}
\textbf{Uniform Sampling:} Uniform Sampling ensures an equal sampling ratio for each method. For example, if initial runs show the highest performance in the order of method A, B, and C, Sampling also follows the order of A, B, and C, then repeats (i.e., A, B, C, A, B, C, ...).
\end{quote}

This sampling process provides a basis for efficient performance enhancements by leveraging a broader search space.

\subsection{Verify answer from sampled runs through LLM Re-ranking}
\label{sec:comparison-experiments}

Previous sampling and voting methods used for maintaining self-consistency (\citealp{zhou2023solving}; \citealp{wang2023selfconsistency}) have the drawback of not fully utilizing the high accuracy upper bound of multiple runs. For example, even if the selection process includes a run that correctly answers previously unsolved problems through improved exploration, sampling and voting tend to favor incorrect answers due to the prevalence of erroneous runs. Since our approach focuses on increasing the search space's upper bound, it is crucial to identify correct answers even from the prevalence of wrong responses. Therefore, we employ LLM re-ranking to derive optimal performance from the selected runs. The re-ranking process follows the methodology proposed by RankGPT \citep{sun2023chatgpt}, which introduces an effective approach for LLM re-ranking.

\section{Experiments}

\paragraph{Setup}

Our experiments are conducted on the subset of MATH dataset \citep{hendrycksmath2021}, which consists of 12,500 challenging math problems from competitions like AMC and AIME, We sampled data from all mathematical domains within the MATH dataset, focusing on questions with difficulty levels 4 and 5. This resulted in 280 challenging questions (comprising approximately 11\% of the entire dataset), which we refer to as MATH-\textit{hard}. We used GPT-4o as the base model for all experiments, and it was also utilized as a LLM re-ranker in Section~\ref{sec:comparison-experiments}. The temperature was set to 0.7 to obtain as diverse responses as possible from each prompting method.

Further details for ablation studies to assess the impact of different components (model size and difficulty level in MATH dataset) can be found in Appendix~\ref{sec:ablation_study}.

\begin{table}[t]
\centering
\resizebox{0.4\textwidth}{!}{%
\begin{tabular}{lcccc}
\toprule
 & \multicolumn{4}{c}{Sampling Methods } \\ 
 \cmidrule(lr){2-5} 
Sample k & Text & Code & CR & \textbf{Uniform} \\
\midrule
base (k=1) & 60.00 & 56.07 & 46.79 & (= Top1) \\ %
\midrule
3 (1,1,1) & 70.0 & 71.07 & 67.5 & 70.71 \\
6 (2,) & 75.71 & 77.5 & 76.07 & 77.14 \\
9 (3,) & 77.86 & 80.36 & 78.57 & \textbf{81.79} \\
12 (4,) & 78.93 & 82.14 & 81.79 & \textbf{84.29} \\
15 (5,) & 79.64 & 82.5 & 82.5 & \textbf{85.36} \\
18 (6,) & 81.79 & 83.21 & 83.21 & \textbf{85.36} \\
21 (7,) & 83.93 & 83.21 & 84.64 & \textbf{86.79} \\
\midrule
Average & 78.27 & 80.00 & 79.18 & \textbf{81.63} \\
\bottomrule
\end{tabular}
}
\caption{\textbf{Search space's upper bound scores on each sampling methods}. Result on MATH-\textit{hard} (* 280 test subset): We increased the number of samples by adopting the default temperature value t=0.7 from CoT-SC. As mentioned in the Method section, each prompting method was based on or reproduced from the following: Text on CoT, Code on CSV (LLM with Local Code Interpreter), and CR from Cumulative Reasoning.}
\label{tab:table1}
\end{table}

\subsection{Efficacy of aggregating distinct prompting methods}
\label{sec:aggregation-experiments}

To quantitatively analyze how effective it is to incorporate various prompting methods, each prompting method was run 21 times, generating 21 different solutions for the entire 280 questions.

This experiment analyzes how the accuracy upper bound changes by incrementally adding runs along the x-axis, comparing the upper bound accuracy of each prompting method against the upper bound obtained through uniform sampling from 21 * 3 runs generated by all prompting methods.

Results from Figure~\ref{fig:exp_1_1} demonstrates our method achieves the highest accuracy of individual prompting methods much earlier; from the 21st to the 11th run for \textit{text}, from the 18th to the 10th run for \textit{Code}, and from the 20th to the 13th run for \textit{CR}, respectively. These results support our hypothesis that employing diverse prompting techniques allows for a more extensive, faster exploration of problems that a single methodology fails to solve or cannot reach.

\begin{table}[t]
\centering
\resizebox{0.5\textwidth}{!}{%
\begin{tabular}{lcccc}
\toprule
 & \multicolumn{4}{c}{Sampling Methods } \\
  \cmidrule(lr){2-5}
Sample k & Text & Code & CR & \textbf{Uniform} \\
\midrule
base (k=1) & 60.00 & 56.07 & 46.79 & (= Top1) \\ %
\midrule
{\textbf{SC (Sample \& Voting)}} \\
3 (1,1,1) & 60.0 & 60.0 & 45.36 & 57.14 \\
6 (2,) & 60.0 & 60.0 & 48.21 & 57.5 \\
9 (3,) & 57.86 & 59.29 & 46.07 & 58.93 \\
12 (4,) & 58.57 & 61.43 & 48.57 & 58.21 \\
15 (5,) & 58.21 & 60.71 & 47.14 & 58.57 \\
18 (6,) & 59.29 & 60.71 & 47.14 & 59.29 \\
21 (7,) & 58.93 & 60.71 & 48.57 & 58.93 \\
  \cmidrule(lr){2-5}
Average & 58.98 & 60.41 & 47.29 & 58.37 \\
\midrule
{\textbf{Rerank@1 (RankGPT, GPT-4o)}} \\
3 (1,1,1) & 63.93 & 63.93 & 60.00 & 62.86 \\
6 (2,) & 64.29	& 66.43 & 65.36 & 65.71 \\
9 (3,) & 64.64 & 68.93 & 66.43 & 64.64 \\
12 (4,) & 65.71 & 69.29 & 67.14 & 66.43 \\
15 (5,) & 65.71 & 71.07 & 67.50 & 66.07 \\
18 (6,) & 65.71 & 71.07 & 67.50 & 66.07 \\
21 (7,) & 65.71 & 71.07 & 68.93 & 65.71 \\
  \cmidrule(lr){2-5}
Average & 65.00 & 68.57 & 66.07 & 65.36 \\
\bottomrule
\end{tabular}
}
\caption{\textbf{Verifying candidate answers} result on MATH-\textit{hard} (* 280 test subset). The experimental settings from Table 1 were maintained, while in Table 2, the verification process for candidate answers found within the search space of each method was performed. The results compare the effectiveness of sample and voting versus LLM Reranking methods. Sampling and voting were performed using Self-Consistency, and LLM Reranking was implemented using RankGPT (GPT-4o, sliding\_window=4, step\_size=2). All accuracy metrics are based on Recall@1.}
\label{tab:table2}
\end{table}

\subsection{Discussion}
The experiments validate our hypothesis that diverse prompting techniques enhance the exploration of the solution space, especially for challenging mathematical problems. By using multiple state-of-the-art prompting methods, we demonstrate that each method explores different parts of the problem space, leading to a more comprehensive and efficient exploration. Consequently, even a simple uniform sampling strategy, when combined with LLM Reranking, results in significant performance improvements and reduced sampling costs. These findings underscore the importance of incorporating multiple methods to achieve a broader and more effective exploration of problem-solving strategies in the MATH domain.

Unfortunately, expanding the search space with multiple agents and using verifiers with a self-consistency algorithm and LLM Reranking do not complement each other. It is due to the foundational philosophy of our algorithm. Our algorithm aims to \emph{collect at least one correct response} by expanding our search space. It does not necessarily mean the correct response appears multiple times.

Therefore, to ensure the final performance improves with the expanded search space, we employed an LLM re-ranking method which is expected to consistently select correct answers, even from sparse values. However, contrary to our expectations, neither the traditional self-consistency (SC) approach nor the LLM re-ranking method consistently guaranteed this improvement.

\section{Conclusion}

In this work we highlight following observations regarding to mathematical reasoning:
\begin{itemize}
    \item Different prompting methods explore distinct solvable problem spaces, and the disparity between these search spaces is challenging to overcome, even by increasing the temperature within a single method.
    \item Therefore, aggregating multiple methods via the sampling approach can expand the solvable problem space, thereby raising the upper bound of accuracy. This approach surpasses the exploration and convergence speed of traditional single-method approaches.
    \item The subsequent LLM re-ranking process yields promising results, demonstrating more efficient approach to produce correct solution compared to the traditional majority voting method used in self-consistency.
\end{itemize}

\section{Limitations}

Our study has yielded insightful findings in the mathematical domain, but it has the following limitations.

\begin{itemize}
  \item Due to the inherent cost issues associated with generating multiple solutions to a single problem, the number of runs produced by each method is not extensive. However, the Appendix~\ref{sec:appendix} describes further experimental results based on GPT-4, where the number of samples was increased to approximately 32\% of the total dataset, compared to the 11\% used in the MATH-\textit{hard} dataset discussed in the main text. These results reaffirm that even with an increased number of runs, differences between output spaces persist when solving difficult problems.
  \item The process of verifying the final answer from sampled runs through LLM re-ranking has shown inconsistent results. Various LLMs (e.g. Gemini 1.5) and methods were tested, but the data did not consistently demonstrate that an increase in the number of runs proportionally enhances both the upper bound of the search space and the final accuracy. It is anticipated that employing a formal math verifier specialized in verification, such as Isabelle\citep{}, as proposed in the DTV paper\citep{}, would ensure that the final accuracy consistently approaches the maximum value of the expanded search space.
  \item We did not incorporate a broader range of problem-solving approaches. Recent studies have introduced promising methodologies for mathematical reasoning, such as agentic prompting methods (e.g. RAT). We leave the evaluation of these diverse methodologies as a future research.
\end{itemize}

\bibliographystyle{acl_natbib}
\bibliography{custom}

\begin{thebibliography}{16}
\expandafter\ifx\csname natexlab\endcsname\relax\def\natexlab#1{#1}\fi

\bibitem[{Chen et~al.(2023)Chen, Ma, Wang, and Cohen}]{chen2023program}
Wenhu Chen, Xueguang Ma, Xinyi Wang, and William~W. Cohen. 2023.
\newblock \href {http://arxiv.org/abs/2211.12588} {Program of thoughts prompting: Disentangling computation from reasoning for numerical reasoning tasks}.

\bibitem[{Claude(2024)}]{claude}
Claude. 2024.
\newblock \href {https://www-cdn.anthropic.com/de8ba9b01c9ab7cbabf5c33b80b7bbc618857627/Model_Card_Claude_3.pdf} {Claude model card}.

\bibitem[{Cobbe et~al.(2021)Cobbe, Kosaraju, Bavarian, Chen, Jun, Kaiser, Plappert, Tworek, Hilton, Nakano, Hesse, and Schulman}]{cobbe2021gsm8k}
Karl Cobbe, Vineet Kosaraju, Mohammad Bavarian, Mark Chen, Heewoo Jun, Lukasz Kaiser, Matthias Plappert, Jerry Tworek, Jacob Hilton, Reiichiro Nakano, Christopher Hesse, and John Schulman. 2021.
\newblock Training verifiers to solve math word problems.
\newblock \emph{arXiv preprint arXiv:2110.14168}.

\bibitem[{Du et~al.(2023)Du, Li, Torralba, Tenenbaum, and Mordatch}]{du2023improving}
Yilun Du, Shuang Li, Antonio Torralba, Joshua~B Tenenbaum, and Igor Mordatch. 2023.
\newblock Improving factuality and reasoning in language models through multiagent debate.
\newblock \emph{arXiv preprint arXiv:2305.14325}.

\bibitem[{Gou et~al.(2024)Gou, Shao, Gong, yelong shen, Yang, Huang, Duan, and Chen}]{gou2024tora}
Zhibin Gou, Zhihong Shao, Yeyun Gong, yelong shen, Yujiu Yang, Minlie Huang, Nan Duan, and Weizhu Chen. 2024.
\newblock \href {https://openreview.net/forum?id=Ep0TtjVoap} {To{RA}: A tool-integrated reasoning agent for mathematical problem solving}.
\newblock In \emph{The Twelfth International Conference on Learning Representations}.

\bibitem[{Hendrycks et~al.(2021)Hendrycks, Burns, Kadavath, Arora, Basart, Tang, Song, and Steinhardt}]{hendrycksmath2021}
Dan Hendrycks, Collin Burns, Saurav Kadavath, Akul Arora, Steven Basart, Eric Tang, Dawn Song, and Jacob Steinhardt. 2021.
\newblock Measuring mathematical problem solving with the math dataset.
\newblock \emph{NeurIPS}.

\bibitem[{Liu et~al.(2023)Liu, Zhang, Li, Liu, and Yang}]{liu2023dynamic}
Zijun Liu, Yanzhe Zhang, Peng Li, Yang Liu, and Diyi Yang. 2023.
\newblock \href {http://arxiv.org/abs/2310.02170} {Dynamic llm-agent network: An llm-agent collaboration framework with agent team optimization}.

\bibitem[{OpenAI(2024)}]{openaiblog}
OpenAI. 2024.
\newblock \href {https://openai.com/index/hello-gpt-4o/} {Gpt-4o blog post}.

\bibitem[{Sun et~al.(2023)Sun, Yan, Ma, Wang, Ren, Chen, Yin, and Ren}]{sun2023chatgpt}
Weiwei Sun, Lingyong Yan, Xinyu Ma, Shuaiqiang Wang, Pengjie Ren, Zhumin Chen, Dawei Yin, and Zhaochun Ren. 2023.
\newblock \href {http://arxiv.org/abs/2304.09542} {Is chatgpt good at search? investigating large language models as re-ranking agents}.

\bibitem[{Wang et~al.(2023)Wang, Wei, Schuurmans, Le, Chi, Narang, Chowdhery, and Zhou}]{wang2023selfconsistency}
Xuezhi Wang, Jason Wei, Dale Schuurmans, Quoc Le, Ed~Chi, Sharan Narang, Aakanksha Chowdhery, and Denny Zhou. 2023.
\newblock \href {http://arxiv.org/abs/2203.11171} {Self-consistency improves chain of thought reasoning in language models}.

\bibitem[{Wei et~al.(2023)Wei, Wang, Schuurmans, Bosma, Ichter, Xia, Chi, Le, and Zhou}]{wei2023chainofthought}
Jason Wei, Xuezhi Wang, Dale Schuurmans, Maarten Bosma, Brian Ichter, Fei Xia, Ed~Chi, Quoc Le, and Denny Zhou. 2023.
\newblock \href {http://arxiv.org/abs/2201.11903} {Chain-of-thought prompting elicits reasoning in large language models}.

\bibitem[{Zhang et~al.(2024)Zhang, Yang, Yuan, and Yao}]{zhang2024cumulative}
Yifan Zhang, Jingqin Yang, Yang Yuan, and Andrew Chi-Chih Yao. 2024.
\newblock \href {http://arxiv.org/abs/2308.04371} {Cumulative reasoning with large language models}.

\bibitem[{Zhong et~al.(2024)Zhong, Wang, and Shang}]{zhong2024debug}
Li~Zhong, Zilong Wang, and Jingbo Shang. 2024.
\newblock \href {http://arxiv.org/abs/2402.16906} {Debug like a human: A large language model debugger via verifying runtime execution step-by-step}.

\bibitem[{Zhou et~al.(2023)Zhou, Wang, Lu, Shi, Luo, Qin, Lu, Jia, Song, Zhan, and Li}]{zhou2023solving}
Aojun Zhou, Ke~Wang, Zimu Lu, Weikang Shi, Sichun Luo, Zipeng Qin, Shaoqing Lu, Anya Jia, Linqi Song, Mingjie Zhan, and Hongsheng Li. 2023.
\newblock \href {http://arxiv.org/abs/2308.07921} {Solving challenging math word problems using gpt-4 code interpreter with code-based self-verification}.

\bibitem[{Zhou et~al.(2024)Zhou, Staats, Li, Szegedy, Weinberger, and Wu}]{zhou2024dont}
Jin~Peng Zhou, Charles Staats, Wenda Li, Christian Szegedy, Kilian~Q. Weinberger, and Yuhuai Wu. 2024.
\newblock \href {http://arxiv.org/abs/2403.18120} {Don't trust: Verify -- grounding llm quantitative reasoning with autoformalization}.

\bibitem[{Zihao et~al.(2024)Zihao, Anji, Haowei, Jiaqi, Xiaojian, and Yitao}]{wang2024rat}
Wang Zihao, Liu Anji, Lin Haowei, Li~Jiaqi, Ma~Xiaojian, and Liang Yitao. 2024.
\newblock Rat: Retrieval augmented thoughts elicit context-aware reasoning in long-horizon generation.
\newblock \emph{arXiv preprint arXiv: 2403.05313}.

\end{thebibliography}

\appendix

\newpage
\section{Appendix}
\label{sec:appendix}

\subsection{Ablation Study}
\label{sec:ablation_study}

\paragraph{Data Sampling Details}
For all MATH data sampling, we fixed random\_seed=42 and adjusted the level, domain, and number of samples to create various data samples.

\begin{quote}
\textbf{a) MATH-hard:}
A subset experimented with GPT-4o in the main text. For hard levels (4 and 5), without domain restrictions (7 domains), 20 samples were drawn each, totaling 280 samples (11.03\% of the test set). This subset, called MATH-hard, allows us to verify reasoning ability on particularly difficult problems.
\end{quote}

\begin{figure}[h] %
    \centering
    \includegraphics[width=0.4\textwidth,keepaspectratio]{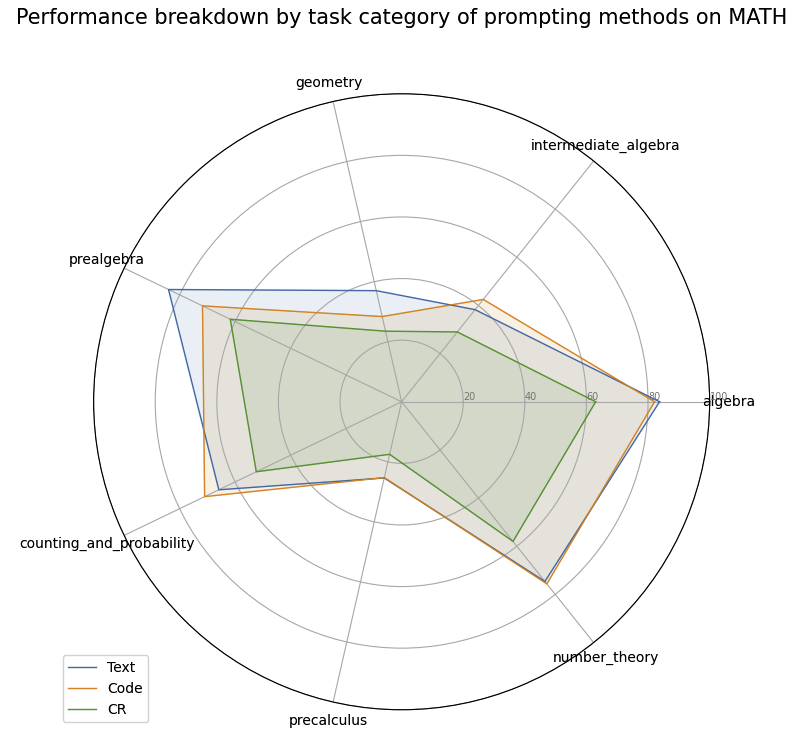}
    \caption{Maximum search space for methods result on MATH-hard (* 280 test subset): Radar graph for showing the average accuracy per all 7 domains for each method (Text, Code, CR) based on their 21 runs.}
    \label{fig:space_agg_20}
\end{figure}

\begin{quote}
\textbf{b) MATH-hard-4doms:}
 Our experimental results showed that even powerful models like GPT-4(o) performed poorly in four specific domains within MATH-hard: "counting\_and\_probability," "geometry," "intermediate\_algebra," and "precalculus" (see Figure~\ref{fig:space_agg_20}). We increased the number of samples in these four domains from 20 to 50, totaling 400 samples (31.55\% of the four domains), creating the MATH-hard-4doms subset.
\end{quote}

\begin{figure}[h] %
    \centering
    \includegraphics[width=0.4\textwidth,keepaspectratio]{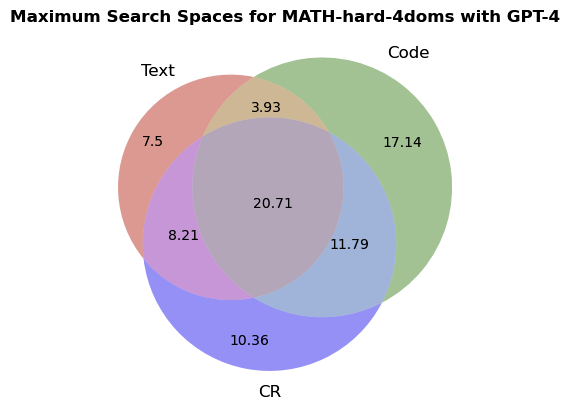}
    \caption{Maximum search space for methods result on MATH-hard-4doms (* 400 test subset): Data sampling details are written in the section above.}
    \label{fig:venn_hard_400}
\end{figure}

\begin{quote}
\textbf{c) MATH-all:}
To verify if the search space expands across the entire set of domains, not just the difficult problems, we sampled 10 samples per domain across all 7 domains and all 5 levels, totaling 350 samples (5\% of the entire dataset).
\end{quote}

\paragraph{Smaller Models on MATH-all}

Previous experiments confirmed that broader approaches are more effective on more difficult problems, leading to the MATH-hard subset for experiments based on GPT-4. As an ablation study, we conducted experiments on MATH-all with general models GPT-3.5-Turbo and LLaMA-3-70B (which performs better than GPT-3.5-Turbo but is similar in cost). We examined whether the search space expands for all levels of problems across each prompting method as the number of method runs samples increases.

\begin{table}[h]
\centering
\resizebox{0.5\textwidth}{!}{%
\begin{tabular}{lcccc}
\toprule
 & \multicolumn{4}{c}{Sampling Methods } \\ 
 \cmidrule(lr){2-5} 
Model: GPT-3.5-Turbo & Text & Code & CR & \textbf{Uniform} \\
\midrule
base (k=1) & 48.86	46.29	42.57 & (= Top1) \\ %
\midrule
5 (2,2,1) & 69.43 & 66.86 & 66.86 & \textbf{70.00} \\
10 (4,3,3) & 76.86 & 78.86 & 73.14 & \textbf{76.57} \\
15 (5,5,5) & 78.86 & 82.29 & 76.29 & \textbf{81.14} \\
20 (7,7,6) & 80.86 & 84.29 & 80.00 & \textbf{84.00} \\
\midrule
Average & 85.36 & 80.07 & 85.86 & \textbf{87.14} \\
\midrule
 & \multicolumn{4}{c}{Sampling Methods } \\ 
 \cmidrule(lr){2-5} 
Model: LLaMA-3-70B & Text & Code & CR & \textbf{Uniform} \\
\midrule
base (k=1) & 65.14 & 41.71 & 61.71 & (= Top1) \\ %
\midrule
5 (2,1,2) & 80.29 & 70.29 & 81.71 & \textbf{82.00} \\
10 (4,3,3) & 85.14 & 80.29 & 85.71 & \textbf{86.86} \\
15 (5,5,5) & 87.43 & 84.29 & 87.14 & \textbf{89.43} \\
20 (7,6,7) & 88.57 & 85.43 & 88.86 & \textbf{90.29} \\
\midrule
Average & 85.36 & 80.07 & 85.86 & \textbf{87.14} \\
\bottomrule
\end{tabular}
}
\caption{\textbf{Search space's upper bound scores on each sampling methods}. Result on MATH-\textit{all} (* 350 test subset): Experimental Details are the same with Table~\ref{tab:table1} and data sampling details are written in the section above.}
\label{tab:table3}
\end{table}

\end{document}